\documentclass[12pt]{article}

\usepackage{newtxtext,newtxmath}

\usepackage{graphicx}
\usepackage{capt-of}

\usepackage[letterpaper,margin=1in]{geometry}

\renewenvironment{abstract}
	{\quotation}
	{\endquotation}

\date{}

\makeatletter
\renewcommand{\fnum@figure}{\textbf{Figure \thefigure}}
\renewcommand{\fnum@table}{\textbf{Table \thetable}}
\makeatother

\usepackage{scicite}

\usepackage{url}

\usepackage{sidecap}

\def\scititle{
Soft Responsive Materials Enhance Humanoid Safety
}
\title{\bfseries \boldmath \scititle}

\author{
	Chunzheng~Wang$^{1\ast}$,
	Yiyuan~Zhang$^{1\ast}$,
	Annan~Tang$^{2\ast}$,
	Ziqiu~Zeng$^{1\ast}$,
    \\
	Haoran~Chen$^{1}$,
	Quan~Gao$^{3}$,
	Zixuan~Zhuang$^{1}$,
	Boyu~Li$^{2}$,
	Zhilin~Xiong$^{2}$,
    \\
	Aoqian~Zhang$^{1}$,
	Ce~Hao$^{1}$,
	Siyuan~Luo$^{1}$,
	Tongyang~Zhao$^{2}$,
    \\
	Cecilia~Laschi$^{1}$,
	Fan~Shi$^{1\dagger}$
    \and
	\small$^{1}$College of Design and Engineering, National University of Singapore, Singapore.\and
	\small$^{2}$ENGINEAI Robotics Technology Co., Ltd, China.\and
	\small$^{3}$Institute of Robotics and Intelligent Systems, ETH Zürich, Switzerland.\and
	\small$^\dagger$Corresponding author. Email: fan.shi@nus.edu.sg\and
	\small$^\ast$These authors contributed equally to this work.
    \thanks{Submitted Date: November 15, 2025}
}

\begin{document} 

\maketitle

\begin{abstract} \bfseries \boldmath


Humanoid robots are envisioned as general-purpose platforms in human-centered environments, yet their deployment is limited by vulnerability to falls and the risks posed by rigid metal–plastic structures to people and surroundings. We introduce a soft–rigid co-design framework that leverages non-Newtonian fluid–based soft responsive materials to enhance humanoid safety. The material remains compliant during normal interaction but rapidly stiffens under impact, absorbing and dissipating fall-induced forces. Physics-based simulations guide protector placement and thickness and enables learning of active fall policies. Applied to a $42$ kg life-size humanoid, the protector markedly reduces peak impact and allows repeated falls without hardware damage, including drops from $3$ m and tumbles down long staircases. Across diverse scenarios, the approach improves robot robustness and environmental safety. By uniting responsive materials, structural co-design, and learning-based control, this work advances interact-safe, industry-ready humanoid robots.
\end{abstract}

\noindent

\section*{INTRODUCTION}


Humanoid robots are envisioned as general-purpose platforms capable of operating seamlessly in human-centered environments~\cite{hirai1998development}. Their anthropomorphic form enables them to use human tools, navigate human spaces, and interact naturally with people. In recent years, rapid advances in actuation, perception, and artificial intelligence have accelerated the deployment of humanoids in manufacturing, logistics, and service settings~\cite{fukuda2017humanoid}. However, as these robots transition from controlled laboratories into the unstructured real world, their safety and robustness have become a primary barrier to widespread adoption. Their tall, rigid structures and inherently unstable bipedal gait make them prone to falling~\cite{subburaman2023survey}. Such failures are not only inevitable but also destructive, posing catastrophic risks to the high-value robot itself, to nearby humans, and to the surrounding environment~\cite{brooks_blog}.

In robotics, researchers have explored various methods to improve the interaction safety. Since most current robots are built from rigid metals or plastics~\cite{morimoto2017soft}, conventional engineering approaches to these safety and durability challenges involve reinforcing rigid structures with stronger materials or increased thickness~\cite{kakiuchi2017development}. However, this engineering approach is fundamentally flawed and creates a vicious cycle: it increases system mass and inertia, which in turn increases the kinetic energy and destructive potential of an impact. Moreover, the increased mass also places greater stress on actuators and reduces energy efficiency. This dilemma drives researchers to refocus on the biomimetic concept of humanoids. The natural compliance of the human musculoskeletal structure helps absorb shocks and distribute impact energy.

Inspired by the compliance of biological systems, researchers have explored a variety of soft-protection strategies for humanoids. One class of approaches employs inflatable bodies to achieve compliance through large deformable volumes \cite{sanan2009robots, best2015control, lee2016active, blumenschein2018helical, alspach2018design}. While effective at cushioning impacts, these systems occupy considerable space, limit range of motion, and suffer from poor heat dissipation, an issue that become critical for high-power actuators. As a result, they are typically applied only to upper-body segments, leaving the lower limbs unprotected. Reactive methods such as deployable airbags have also been proposed \cite{kajita2016impact}, but these devices are single-use, costly, and provide coverage only in specific regions or directions. Together, these limitations underscore the difficulty of achieving compact, reusable, and comprehensive protection for life-size humanoids.

Despite significant advances in soft robotics \cite{pfeifer2012challenges, laschi2016soft}, fully soft humanoids remain impractical at human scale. Existing soft-legged robots are mostly small prototypes \cite{xia2021legged, drotman20173d, wu2019insect, bern2019trajectory}, as large compliant structures face challenges in maintaining strength, load capacity, precise control, and thermal stability required for life-size bipedal locomotion~\cite{hawkes2021hard}. Even in nature, biological organisms rely on soft–rigid hybrids: rigid bones provide structural support, while compliant tissues absorb shocks and distribute loads~\cite{delp2007opensim}. The same principle applies to humanoid robots, where rigid skeletons are indispensable for mechanical strength, but can be complemented by soft, adaptive materials to ensure safety during interaction~\cite{morimoto2017soft}. Moreover, most current humanoid platforms already benefit from mature rigid-robot ecosystems, including standardized hardware, simulation tools, and control frameworks~\cite{IEEE_Humanoid_Report_2025}. Replacing these with entirely soft alternatives would be costly and disruptive. A more practical and scalable approach is to pursue soft–rigid hybrid integration, which enhances safety and resilience while leveraging the strengths of existing infrastructure~\cite{asano2017design}.


Here, we identify non-Newtonian fluid–based soft responsive materials (SRMs)~\cite{barnes1989introduction} as an ideal candidate for achieving soft–rigid hybrid protection~\cite{yang2025hybrid}. SRMs are lightweight, compact, reusable, and highly deformable materials that can nevertheless absorb large impulsive loads with remarkable efficiency. Their unique combination of flexibility and impact protection has led to their widespread use in extreme-sports protective gear~\cite{D3O, RHEON, AiroFoam}. As a class of responsive polymers, they exhibit shear-thickening behavior~\cite{lee2003ballistic}, in which viscosity increases dramatically under high strain rates. They remain soft and flexible under normal operating conditions, maintaining behavior close to non-Newtonian properties. However, upon sudden impact, they undergo a rapid and reversible solid-like transition on a sub-millisecond timescale~\cite{waitukaitis2012impact}. It enables efficient dissipation of impact energy and provides condition-selective protection that conventional materials cannot achieve~\cite{zhao2020shear}. In our preliminary tests, even a $3$ mm SRM layer was sufficient to protect a fragile porcelain plate from a $2.5$ kg impact dropped from $60$ cm (Movie 1), highlighting the material’s strong protective performance.

In this work, we present a soft–rigid co-design framework that integrates SRMs with life-size humanoid robots to enhance both structural and interaction safety. The SRM forms a soft, reusable protective layer applied to critical joints and surface regions, remaining compliant during normal operation but stiffening instantly upon impact. A computational design pipeline optimizes the placement and thickness of these materials, while a learning-based control policy is trained to actively exploit their passive energy-dissipating properties, directing impact forces toward protected zones. The proposed framework is platform-agnostic and applicable to any standard humanoid platform. 

We evaluate the framework through three integrated phases of development and testing. First, we characterize the impact-response behavior of the proposed SRM composites, showing more effective impulse absorption compared with conventional protective materials. Second, we assess the protection performance on joint modules with one to three degrees of freedom, where the SRM layers significantly reduce peak impact pressure in repeated drop tests, extending hardware durability while preventing damage to surrounding surfaces and objects. Finally, through large-scale simulations and real-world fall experiments, we identify the regions of the humanoid most prone to damage and apply the protective layers accordingly. The robot withstands diverse fall conditions, including multi-level drops, stair descents, outdoor uneven terrain, and dynamic motions such as running, turning, and front flipping, while remaining fully operational. A learning-based active fall policy is further trained to use the protected areas effectively. Together, these results establish a unified material–control paradigm that advances humanoid robots toward safer, more resilient, and real-world-ready operation in human-centered environments.

\section*{RESULTS}

\begin{figure} 
	\centering
	\includegraphics[width=1\textwidth]{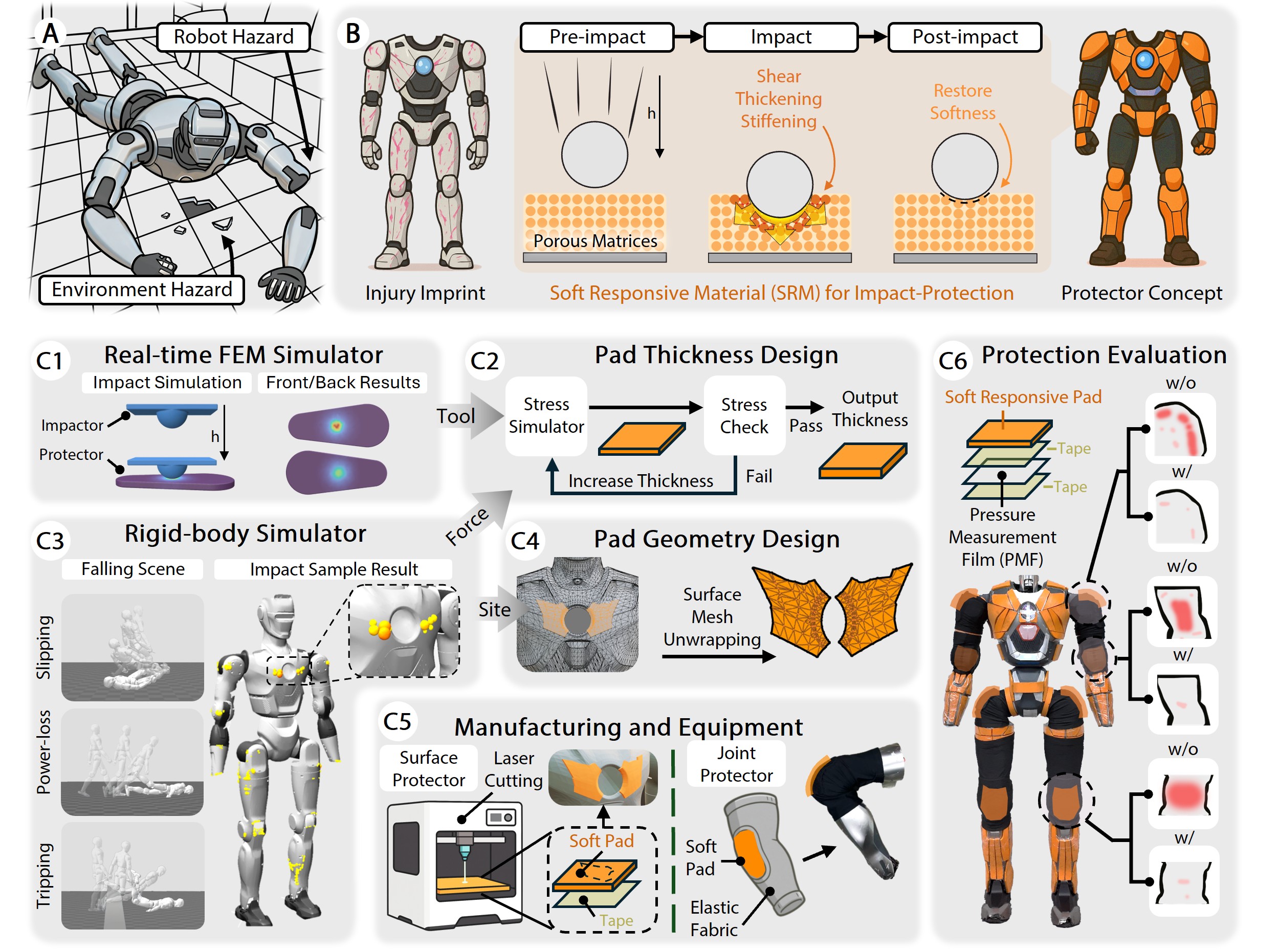} 

	\caption{
    \textbf{Soft responsive material protectors enhance humanoid robot safety.}
    (A) Typical safety hazards associated with humanoid robots during falls and collisions.
    (B) Soft responsive materials (SRMs) provide effective impact absorption.
    (C1) Real-time finite-element method (RT-FEM) simulation of impact response.
    (C2) Protector thickness optimization based on RT-FEM analysis.
    (C3) Large-scale humanoid impact sampling using physics simulation.
    (C4) UV-mapping workflow for unfolding 3D robot surfaces into 2D templates for protector design.
    (C5) Fabrication of surface and joint protectors through monolithic laser cutting.
    (C6) Full-body humanoid experiments validating the effectiveness of SRM impact protection.
    }
	\label{fig:pipeline} 
\end{figure}


\subsection*{Overview of co-design and validation phases}

Humanoid robots pose safety challenges because their large mass, rigid structures, and unstable bipedal gait can generate large impact forces on both the robot and its surrounding environment (Fig.~\ref{fig:pipeline}A). Our overarching hypothesis is that a compact, surface-mounted layer of non-Newtonian, soft responsive material (SRM) can significantly mitigate these impacts, providing simultaneous protection to the hardware and the environment.


Fig.~\ref{fig:pipeline} and Movie~S1 summarize the workflow and key findings. The results follow a structured, three-phase progression, from material characterization, to joint-level validation, and finally to whole-body humanoid testing, demonstrating how soft–rigid co-design scales from local impact damping to full-system safety enhancement.

Phase I – Material characterization: 
European Conformity (CE) impact tests confirmed that the SRM absorbs shock loads far more effectively than conventional protective materials (Fig.~\ref{fig:pipeline}B). Force–time measurements were used to calibrate a real-time finite element model (RT-FEM) within our simulation framework (Fig.~\ref{fig:pipeline}C1). The simulated pressure distributions closely matched the experimental measurements, validating the ability of the model to reproduce the material’s impact-damping behavior and enabling accurate prediction for subsequent co-design stages in Fig.~\ref{fig:pipeline} (C2 and C3).

Phase II - Joint-level protection: 
SRM patches were automatically co-designed, UV-mapped, and laser-cut to conform tightly to the curved geometry of 1-, 2-, and 3-DoF humanoid joints without restricting their motion range as Fig.~\ref{fig:pipeline} (C4 and C5). Drop and impact tests demonstrated that the SRM layers significantly reduced peak pressure and increased the number of safe consecutive falls compared with unprotected joints.

Phase III – Whole-body validation: 
Life-size humanoid experiments across diverse fall scenarios, including stair descents, multi-level drops, and dynamic flipping impacts, showed that robots equipped with the SRM protection remained fully operational after repeated impacts. Even under severe conditions, the protector markedly reduced both structural damage and environmental impact. Pressure measurement films (PMFs) attached to the robot surface quantitatively confirmed significant reductions in peak impact pressure at key contact regions (Fig.~\ref{fig:pipeline} C6).

\subsection*{Impulse absorption of soft responsive materials}



\begin{figure}[!htbp]
    \centering
	\includegraphics[width=0.8\textwidth]{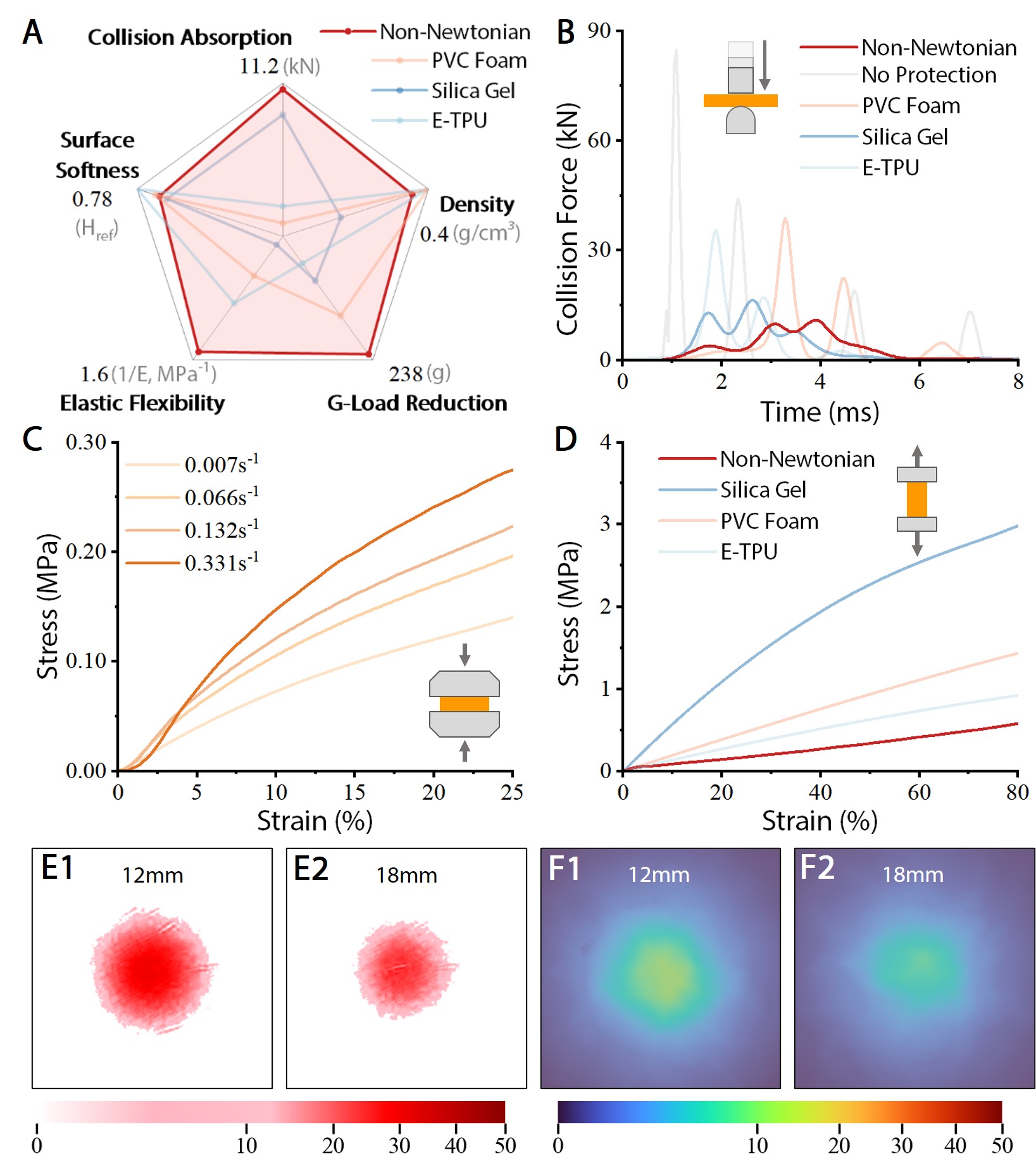} 
    \caption{\textbf{SRM mechanical characterization and RT-FEM simulation.}
        (A) Comparison of soft responsive material (SRM) with conventional protective materials, demonstrating its superior impact absorption and mechanical performance.
        (B) High-frequency impact response profiles of SRM versus common protective materials.
        (C) Strain-hardening behavior of SRM under increasing compression rate, demonstrating its strain-rate-dependent stiffening.
        (D) Tensile properties of SRM compared with other protective materials, showing higher stretchability and elastic flexibility.
        (E1–E2) PMF measurements showing impact distributions with 12 mm and 18 mm SRM pads.
        (F1–F2) Corresponding RT-FEM results, demonstrating the accuracy of our simulation.}
    \label{fig:material}
\end{figure}

The impact absorption performance of the proposed soft responsive material (SRM) was benchmarked against commonly used industrial protective materials, including PVC foam, silica gel, and expanded thermoplastic polyurethane (E-TPU), as shown in Fig.~\ref{fig:material} (A and B). Following the European CE certification protocol~\cite{EN1621_1_2012}, an $5 \mathrm{kg}$ payload was dropped from a height of $1 \mathrm{m}$ onto each sample. Contact forces were recorded using a CHENGTEC CT3105 high-range force sensor ($0$–$100$ kN, sensitivity $4$ pC/N)~\cite{chengtec} paired with a high-speed Measurement Computing data acquisition card sampling at $25$ kHz~\cite{mc-daq}.

A $6 \mathrm{mm}$ SRM layer reduced peak impact force by  $70.0\%$ from $86.8 \mathrm{kN}$ to $25.8 \mathrm{kN}$, while a $12 \mathrm{mm}$ layer achieved an $87.1\%$, lowering the peak to $11.2 \mathrm{kN}$. Beyond lowering the peak magnitude, the SRM significantly altered the shape of the force–time curve: compared with other materials, it produced a longer impulse duration, stronger damping, and much lower rebound force (Fig.~\ref{fig:material}B). These characteristics reflect more efficient impulse dissipation, which is an essential requirement for protecting humanoid joints and link structures from impulsive loading.

Slow-motion recordings (Movie~S1) provide visual confirmation of these effects: the impacting object rapidly settles on the SRM surface with no observable bounce, whereas conventional foams and elastomers exhibit repeated bouncing. These qualitative observations match the high-frequency force data, and highlights the SRM’s strong damping and impulse dissipation.

The material’s underlying mechanism is illustrated in Fig.~\ref{fig:material} (C and D). As the compression rate increases, the slope of the stress–strain curve (effective modulus) rises sharper, demonstrating its strain-rate-dependent stiffening. Tensile tests further reveal the SRM’s large elastic deformation capacity, indicating that the material can flexibly conform to curved robot surfaces while still delivering high impact resistance.

Conventional materials, especially thick rubber or Silica Gel, require roughly twice the thickness in our tests to achieve comparable peak-force attenuation, and their stiffness limits their suitability for protecting complex humanoid joint geometries~\cite{zinn2004new}. In contrast, the SRM offers high impact performance in a thin, lightweight, and deformable form factor, making it well suited for integration onto articulated, space-constrained humanoid robot structures.

\subsection*{Simulation-based calibration and modeling of SRM impact behavior}

To accurately capture the impact response of the non-Newtonian SRM in simulation, we adopt a phenomenological modeling approach based on simulation-based calibration. Rather than modeling particle-scale shear-thickening behavior, we approximate the SRM as a rate-dependent elastic material and calibrate this model directly from the measured force–time curves obtained in the CE drop-weight tests (Fig.~\ref{fig:material}B). The model is implemented within our unified real-time elastic simulation framework~\cite{zeng2025fba}, allowing the material parameters to be adjusted so that the simulated impact profiles match experimental measurements over the full temporal evolution of the contact event.

This calibration process constrains the effective stiffness, damping, and rate sensitivity of the SRM, enabling the simulator to reproduce the material’s rapid stiffening under high strain rates without explicitly modeling its underlying colloidal microstructure. Because the procedure fits the entire force–time trajectory, rather than isolated peak values, the resulting model robustly captures the SRM’s dynamic mechanical behavior, including its impulse absorption, energy dissipation, and time-dependent load redistribution, throughout the full duration of impact.

Using the $12$ mm material drop tests as ground truth, the calibrated simulation accurately predicts peak pressures across other thicknesses, and the resulting pressure fields closely match those measured using pressure measurement films (Fig.~\ref{fig:material}E1 and Fig.~\ref{fig:material}F1). For a previously unseen $18$ mm sample, the simulator also reproduces the measured pressure pattern (Fig.~\ref{fig:material}E2 and Fig.~\ref{fig:material}F2), further validating that the simplified rate-dependent elastic model captures the essential load-spreading and energy-dissipation behavior of the SRM (Movie~S2). This level of fidelity is sufficient for determining protector placement and thickness  (Fig.~\ref{fig:supp-sim-results}) during later co-design stages.

Beyond matching experiments, the calibrated model provides useful physical insights. It shows that the macroscopic impact response of the SRM can be captured without micro-scale simulation, enabling efficient and scalable modeling of humanoid-scale falls. The alignment between simulation and experiment suggests that rate-dependent elasticity is the dominant factor governing SRM performance at sub-millisecond-scale impacts. Finally, because the calibrated model runs in real time or near–real time, it supports rapid design iteration, prediction of pressure hotspots under diverse fall scenarios, and high-throughput simulation-in-the-loop co-design.

\subsection*{Joint modules protection through soft–rigid co-design}

\begin{SCfigure}[1.0]
    \centering
	\includegraphics[width=0.6\textwidth]{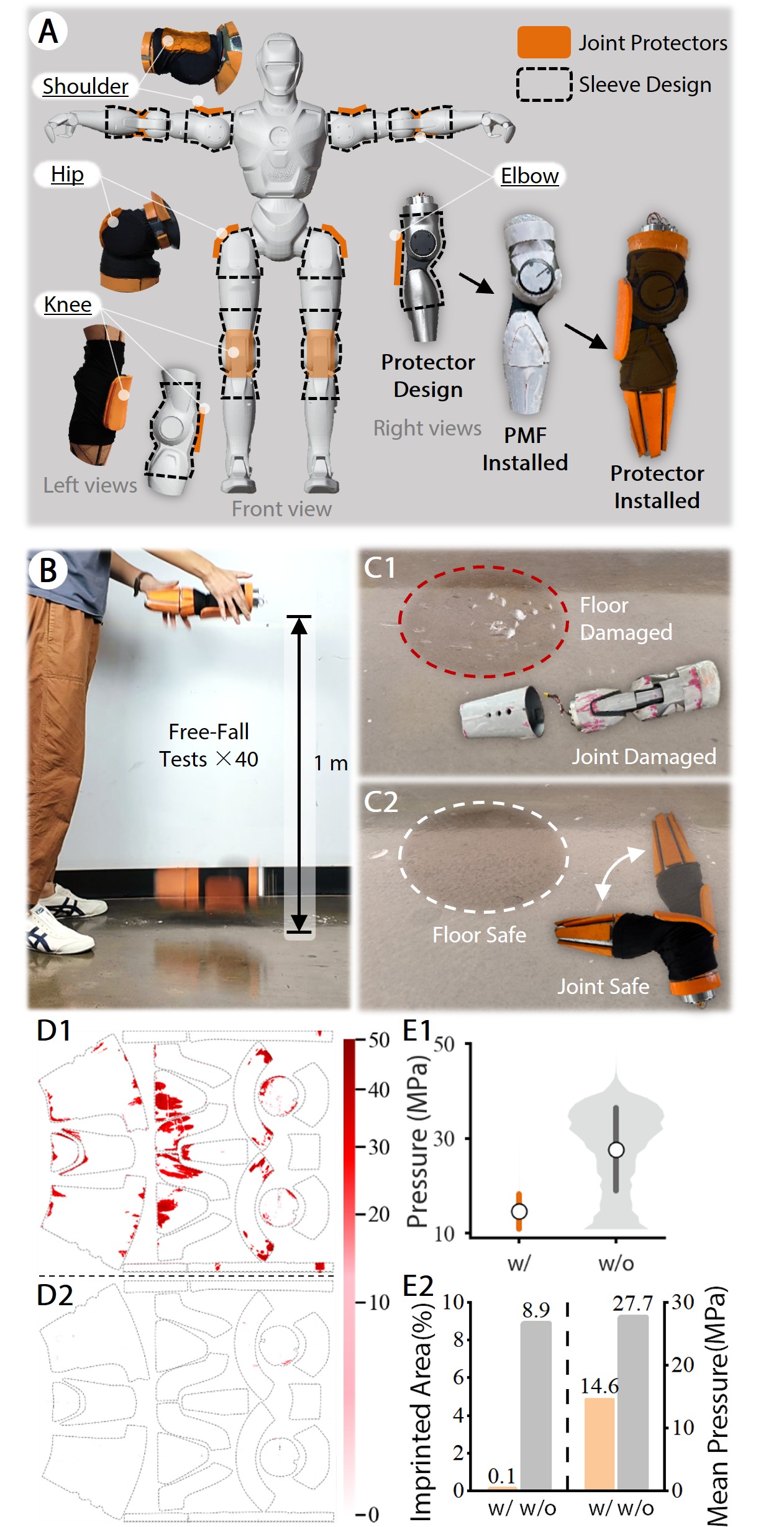} 
    \caption{\textbf{Joint protection of humanoid robots and impact-protection evaluation.}
    (A) Protective-pad designs equipped with PMFs for different humanoid joint modules.
    (B) Forty-drop impact test of the elbow joint onto a concrete surface.
    (C1) Unprotected elbow joint: gear jamming and structural fracture occur, accompanied by visible damage to the concrete surface.
    (C2) Protected elbow joint: normal joint rotation is preserved and no surface damage is observed.
    (D1) PMF (2D unfolded) pressure distribution for the unprotected joint.
    (D2) PMF (2D unfolded) pressure distribution for the protected joint.
    (E1) Violin plot showing the full distribution and median of impact pressures with and without the protector.
    (E2) Comparison of the area and mean pressure of regions exceeding $10$ MPa with and without the protector.
    }
    \label{fig:joint-design}
\end{SCfigure}

Humanoid robots contain three primary types of joint modules, including 1-DoF joints such as elbows and knees, 2-DoF joints such as shoulders, and 3-DoF joints such as hips, which collectively bear the majority of impact loads during falls as Fig.~\ref{fig:joint-design} (A and B). To evaluate the effectiveness of the proposed soft–rigid co-design, we integrated the SRM-based protective layer onto representative joints in each category, shaped through UV-mapping~\cite{Levy2002LSCM} to conform tightly to the curved geometries without restricting mobility (Fig.~\ref{fig:mujuco-sample}A).

Across all joint modules, the improvement in durability was significant. Unprotected joints consistently failed after around twenty consecutive drops from a height of $1 \mathrm{m}$ (Fig.~\ref{fig:joint-design}B), exhibiting structure deformation or gear jamming (Movie~S3). After forty falling, outer shells are cracked and impacts also caused visible damage to the floor surface (Fig.~\ref{fig:joint-design}C1), indicating high localized stresses during contact. In contrast, joints equipped with the SRM protector remained fully functional after more than forty consecutive impacts, showing no loss of actuation performance, no internal damage, and no observable damage to the floor (Fig.~\ref{fig:joint-design}.C2).

Quantitative measurements using PMFs further support these observations. Compared with unprotected joints (Fig.~\ref{fig:joint-design}D1), the protected joints exhibited sharply reduced contact pressures, with high-stress regions almost completely eliminated (Fig.~\ref{fig:joint-design}D2). With the SRM protector in place, most impact pressures remained below 
$10$ MPa, appearing as white or near-white on the PMF due to the film’s lower detection threshold (see Supplementary Material: Pressure measurement film). The overall pressure distribution shifted markedly downward (Fig.~\ref{fig:joint-design}E1), and the area of regions exceeding $10$ MPa was significantly reduced (Fig.~\ref{fig:joint-design}E2). This drastic reduction in localized pressure aligns with the absence of structural damage on the protected joints and the lack of floor scarring or cracking during repeated tests.

The pressure maps also highlight the stress reduction on housings, bearing seats, and mounting plates. This benefit is particularly significant for multi-DoF joints, where complex curved geometries make rigid armor difficult to apply uniformly. Despite the added protection, the SRM layer remains thin and lightweight, causing negligible increases in inertia and preserving the joints’ full range of motion. 

Together, these results show that joint-level soft–rigid co-design provides an effective structural buffer between the robot and its environment. It dramatically improves fall tolerance and durability while maintaining mobility and dynamic performance, the capability essential for safe humanoid operation in real-world environments.

\subsection*{Damage mapping: simulation-based testing vs. real-world failures}

\begin{figure} [!htbp]
	\centering
    \vspace{-15mm}
	\includegraphics[width=0.95\textwidth]{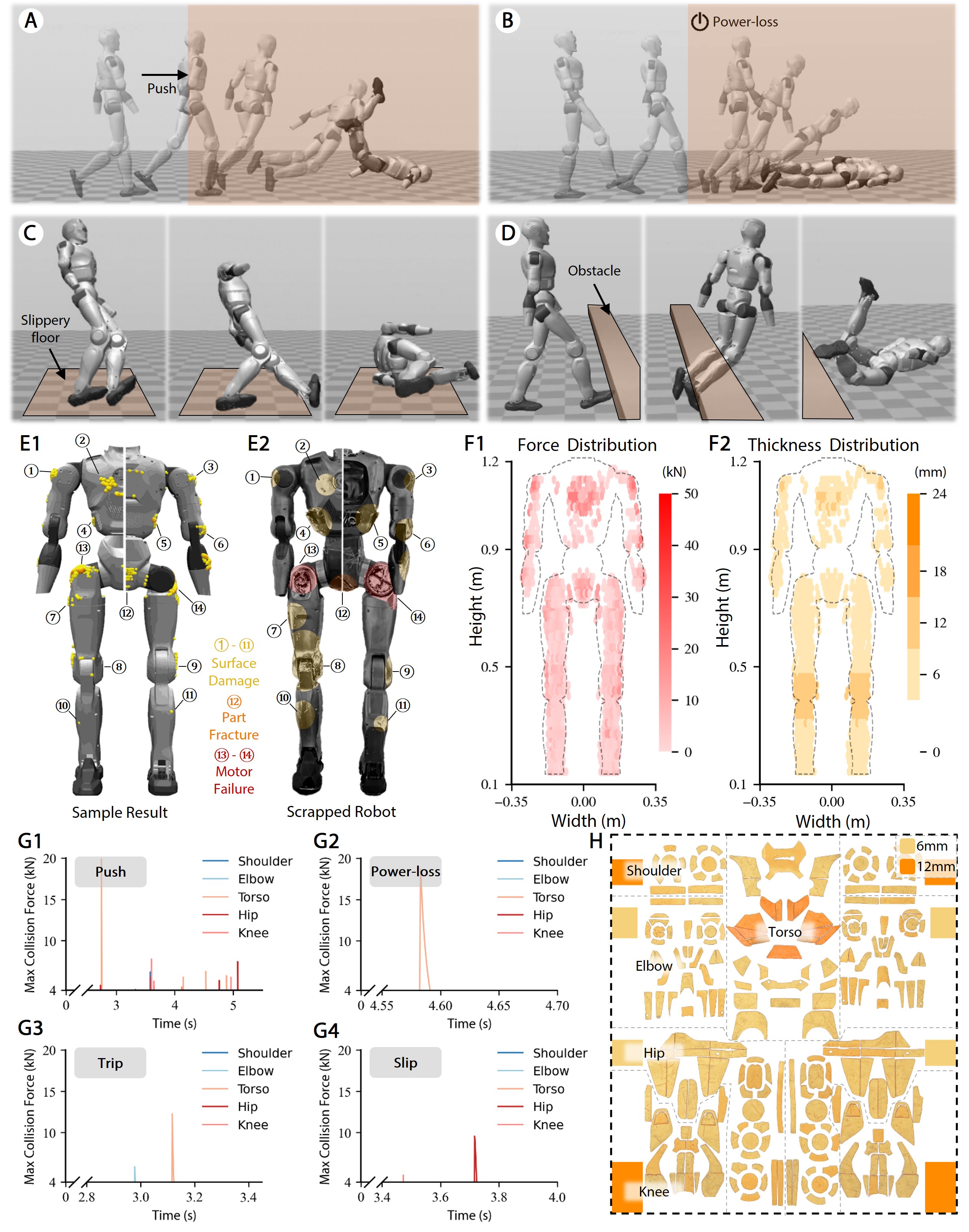} 

    \vspace{-8mm}
	\caption{\textbf{Large-scale simulation of common humanoid failure modes and resulting impact analysis.}
    (A) External push-over.
    (B) Unexpected power loss.
    (C) Ground-slip fall.
    (D) Obstacle-induced trip.
    (E1–E2) Aggregated impact-location statistics across all simulated failure modes, compared with damage patterns from scrapped humanoid robots.
    (F1–F2) Resulting whole-body force distribution and the corresponding protector placement and thickness determined by co-design.
    (G) Time-resolved impact-force profiles for major body segments during representative falls.
    (H) Final protective-pad layout produced by the co-design pipeline.}
	\label{fig:mujuco-sample} 
\end{figure}


To determine optimal material thickness and placement, we conducted large-scale simulation falling experiments covering four representative falling scenarios (Movie~S4): external pushing (400 trials), electrical shutdown (400), obstacle collision (400), and slipping (100) as shown in Fig.~\ref{fig:mujuco-sample} (A–D). For each case, we recorded the spatial distribution and magnitude of contact forces over multiple episodes to build a statistical “damage map” of the humanoid body. The aggregated results revealed that impacts are highly non-uniform, concentrating on specific regions such as the outer limbs, hips, and torso panels (highlighted in yellow in Fig.~\ref{fig:mujuco-sample}E).

To validate these predictions, we compared the simulated contact distributions with physical evidence from previously damaged humanoid platforms after long tests. The correspondence between simulated hotspots and real failure locations confirms that the simulation reasonably captures the dominant stress pathways during typical fall events (Fig.~\ref{fig:mujuco-sample}F1). This consistency supports the use of simulation-based analysis to identify vulnerable regions and to guide the targeted placement of the SRM protectors (Fig.~\ref{fig:mujuco-sample}F2).

Beyond validation, the damage mapping also provides practical design insights. It shows that damage on humanoids occurs mainly on the sides of the body, where lateral falls produce concentrated impact loads. High-stress regions also appear near major joints, which tend to protrude and make early contact during a fall (Fig.~\ref{fig:mujuco-sample}G). These findings suggest that protection should focus on these vulnerable areas rather than applying uniform padding (Fig.~\ref{fig:mujuco-sample}H), allowing the design to improve safety while keeping the robot lightweight and maintaining joint mobility.

\subsection*{Whole-body passive fall robustness enabled by SRM protection}

\begin{figure} 
	\centering
	\includegraphics[width=1\textwidth]{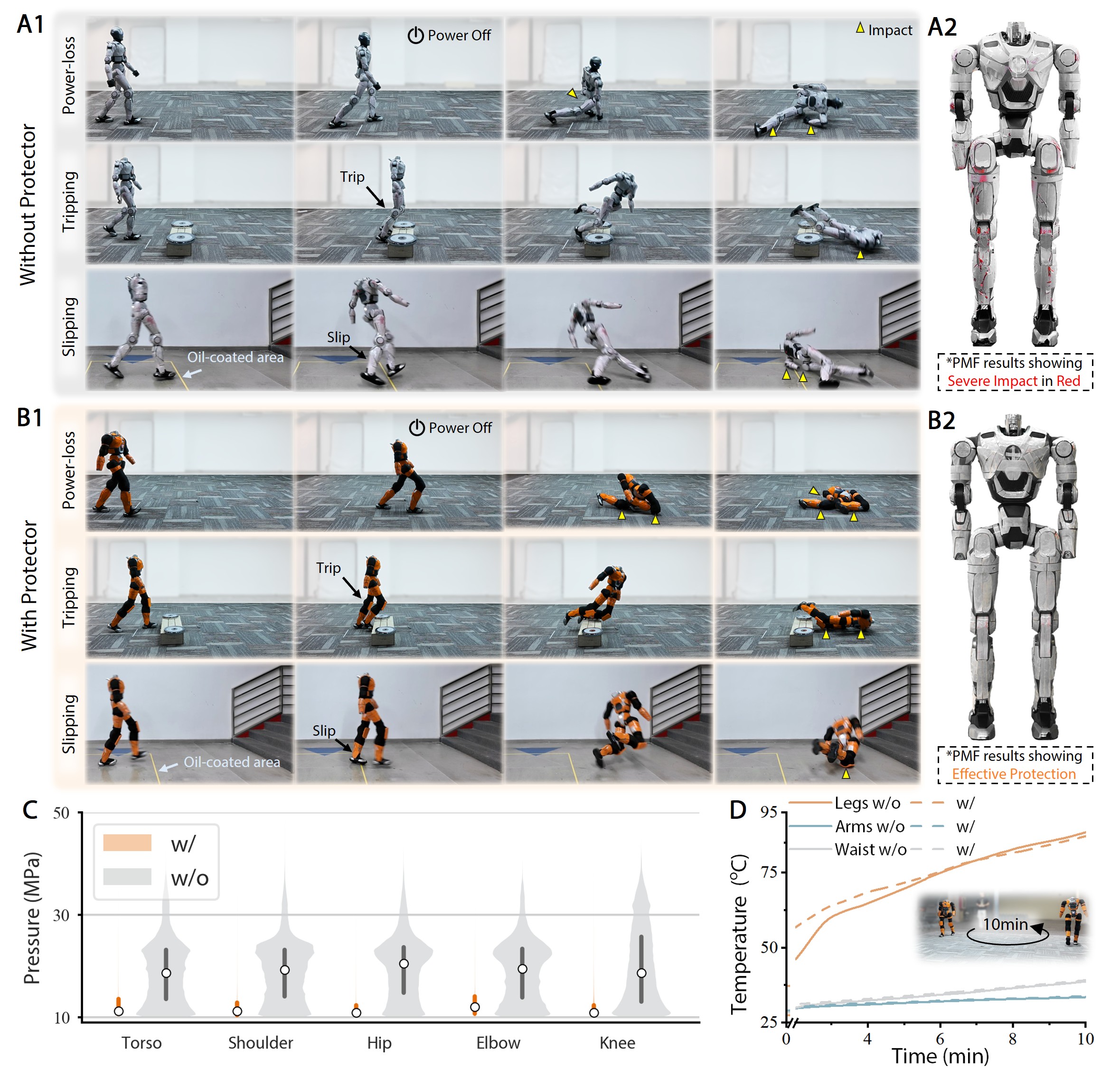} 

	\caption{\textbf{Physical comparison experiments across failure modes with and without protectors.}
        (A1) Power-loss, tripping, and slipping tests conducted without protectors, with yellow triangular markers indicating ground-impact locations.
        (A2) PMF results after repeated unprotected tests, showing extensive high-pressure regions (red traces) across the robot surface.
        (B1) Power-loss, tripping, and slipping tests conducted with SRM protectors installed.
        (B2) PMF results after repeated protected tests, exhibiting minimal high-pressure markings.
        (C) Violin-plot comparison of impact-intensity distributions with and without protectors.
        (D) Motor-temperature evaluation of the humanoid robot in protected and unprotected conditions.}
	\label{fig:pressure-compare} 
\end{figure}


We conducted extensive real-world fall experiments to evaluate the passive safety of the PM01 humanoid robot (see Supplementary Material under "Humanoid robot platform" section) under representative failure modes, including slipping, power loss, and obstacle-induced falls as Fig.~\ref{fig:pressure-compare} (A1 and B1). In the unprotected condition, the robot suffered severe hardware failures after only one or two dynamic falls, such as sideways slips, including cracked structural housings, damaged circuit boards, and jammed or broken actuators. In contrast, with the SRM-based protectors installed, the robot completed more than thirty consecutive falls and recoveries without visible damage or loss of functionality, demonstrating a pronounced increase in impact tolerance (Movie~S5).

Pressure measurements further quantified the improvement in protective performance as Fig.~\ref{fig:pressure-compare} (A2 and B2). With the SRM installed, PMF results showed a clear shift from concentrated red, high-pressure peaks to broadly distributed low-pressure regions, with most contact areas falling below $10$ MPa (Fig.~\ref{fig:pressure-compare}C). The area of the robot surface experiencing pressures above $10$ MPa dropped dramatically from 
$674.8$ cm$^2$ to $142.6$ cm$^2$, accounting for only $1.3\%$ of the total robot surface area. In addition, the median impact pressure decreased from $19.5$ MPa to $11.3$ MPa, corresponding to the $42.1\%$ reduction. These quantitative improvements confirm that the SRM protector suppresses high-pressure spikes and redistributes impact loads over safer, lower-stress regions.

Despite adding a new protective layer, the robot’s operational characteristics remained unchanged. Reachability was fully preserved, and motor temperatures showed no noticeable increase after ten minutes of continuous walking (Fig.~\ref{fig:pressure-compare}D), indicating that the SRM introduces minimal thermal or mechanical side effects. These results confirm that the protector provides significant safety benefits without compromising mobility or system performance

Robustness was further assessed through dynamic and high-impact stress tests, including front flipping, dancing, running, walking, rolling down long staircases, and controlled drops from a second-story platform as Fig.~\ref{fig:real-highlight} (A–F). Across all trials, the humanoid maintained full functionality with no structural failures or actuation loss (Movie~S7). These results confirm that the proposed SRM protector enables life-size humanoid robots to survive and operate reliably under impacts that would normally result in irreversible hardware damage, thereby significantly enhancing passive fall resilience in real-world environments.

\subsection*{Environmental impact mitigation enabled by SRM protection}




\begin{figure}
    \centering
	\includegraphics[width=1.0\textwidth]{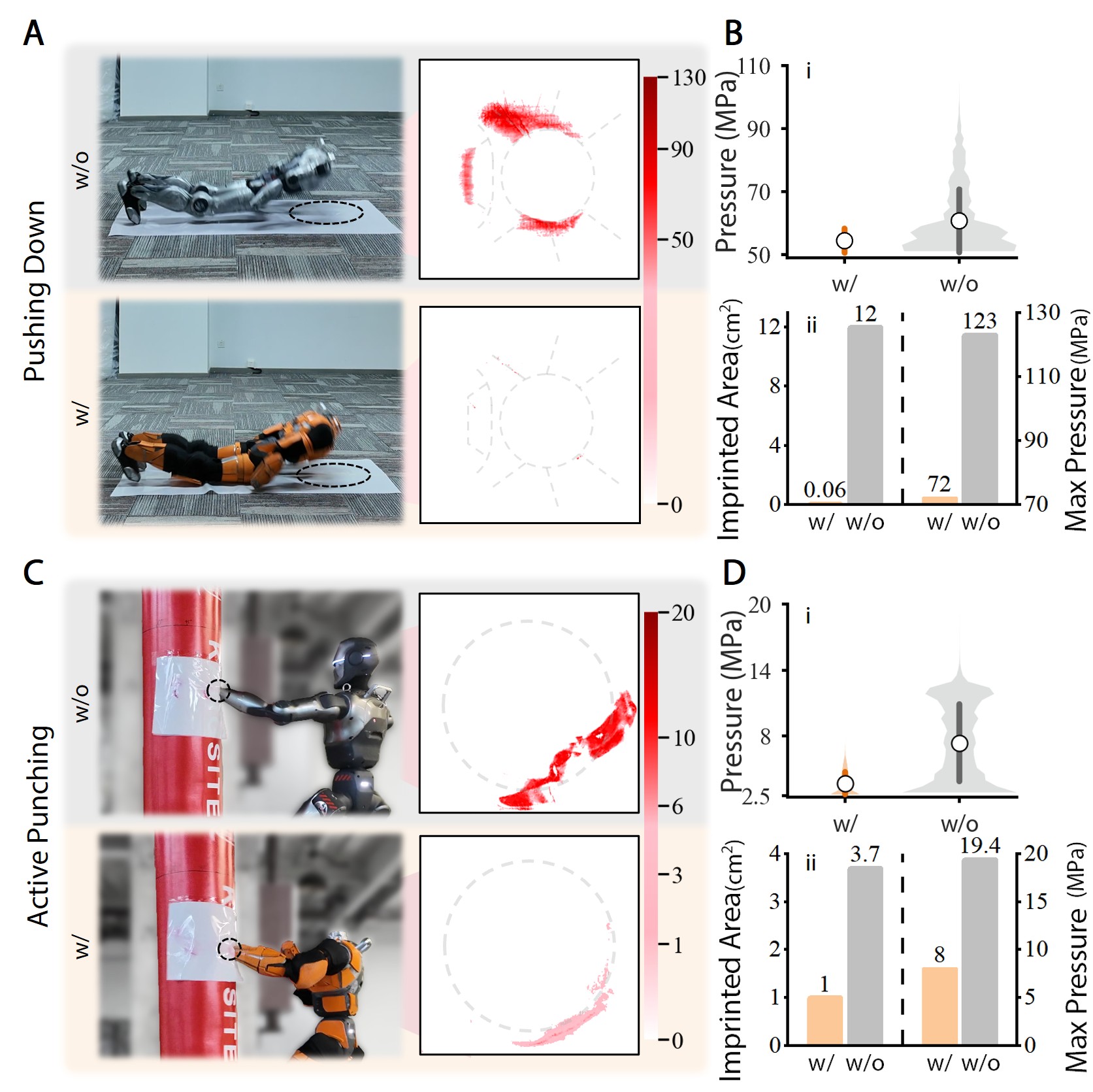} 
    \caption{\textbf{Protectors reduce the robot’s risk of damaging the environment.}
    (A) Flat-ground push-down tests with and without protectors. PMFs attached to the floor capture the torso–ground impact region.
    (B) Punching-bag impact tests with and without protectors. PMFs attached to the bag capture the arm–bag impact region.
    (C) Pressure distributions, high-pressure area fractions, and maximum pressures for the push-down tests.
    (D) Corresponding statistics for the punching-bag tests.
    Notably, high-pressure PMFs were used for the push-down tests, while low-pressure PMFs were used for the punching-bag tests.
    }
    \label{fig:env-protect}
\end{figure}

In addition to improving robot robustness, the proposed soft–rigid design also enhances safety for the surrounding environment and any objects or humans that may be in contact during a fall. During repeated drop tests of joint modules, unprotected joints produced clear dents and surface fractures on the test platform after approximately forty impacts (Fig.~\ref{fig:joint-design}C1). In contrast, the same tests with the SRM protector produced no observable surface damage (Fig.~\ref{fig:joint-design}C2), indicating that the deformable layer effectively dissipates contact energy before it transfers to the environment.

To quantify environmental protection at whole-body scale, we performed controlled push-down tests with PMF sheets placed beneath the humanoid’s torso (Fig.~\ref{fig:env-protect}A). Under identical falling postures, the SRM-equipped robot produced substantially lower ground pressure. The maximum recorded pressure dropped from $122.8$ MPa to $71.6$ MPa ($42.7\%$ reduction), and the area exceeding $50.0$ MPa (the lower bound of the high-pressure PMF range) shrank from $12.0$ cm$^2$ to just $0.06$ cm$^2$ (Fig.~\ref{fig:env-protect}B). Most regions registered within the film’s low-pressure range, indicating broad load distribution rather than concentrated impact points. 

We further evaluated interaction safety with external objects using a sandbag instrumented with PMF (Fig.~\ref{fig:env-protect}C). When the humanoid’s arm struck the bag at a fixed velocity and position, the SRM protector again reduced impact severity: maximum pressure decreased from $19.4$ MPa to $8.0$ MPa ($58.8\%$ reduction), and the area exceeding $2.5$ MPa (the lower bound of the low-pressure PMF range) decreased from $3.7$ cm$^2$ to $1.0$ cm$^2$ (Fig.~\ref{fig:env-protect}D). 

Together, these results demonstrate that the soft–rigid design not only protects the robot’s hardware but also substantially reduces the impact forces transmitted to the surrounding environment (Movie~S6), an essential requirement for safe, human-compatible operation in shared spaces.

\subsection*{Learning-based policies for active fall protection}

\begin{figure} 
	\centering
	\includegraphics[width=1\textwidth]{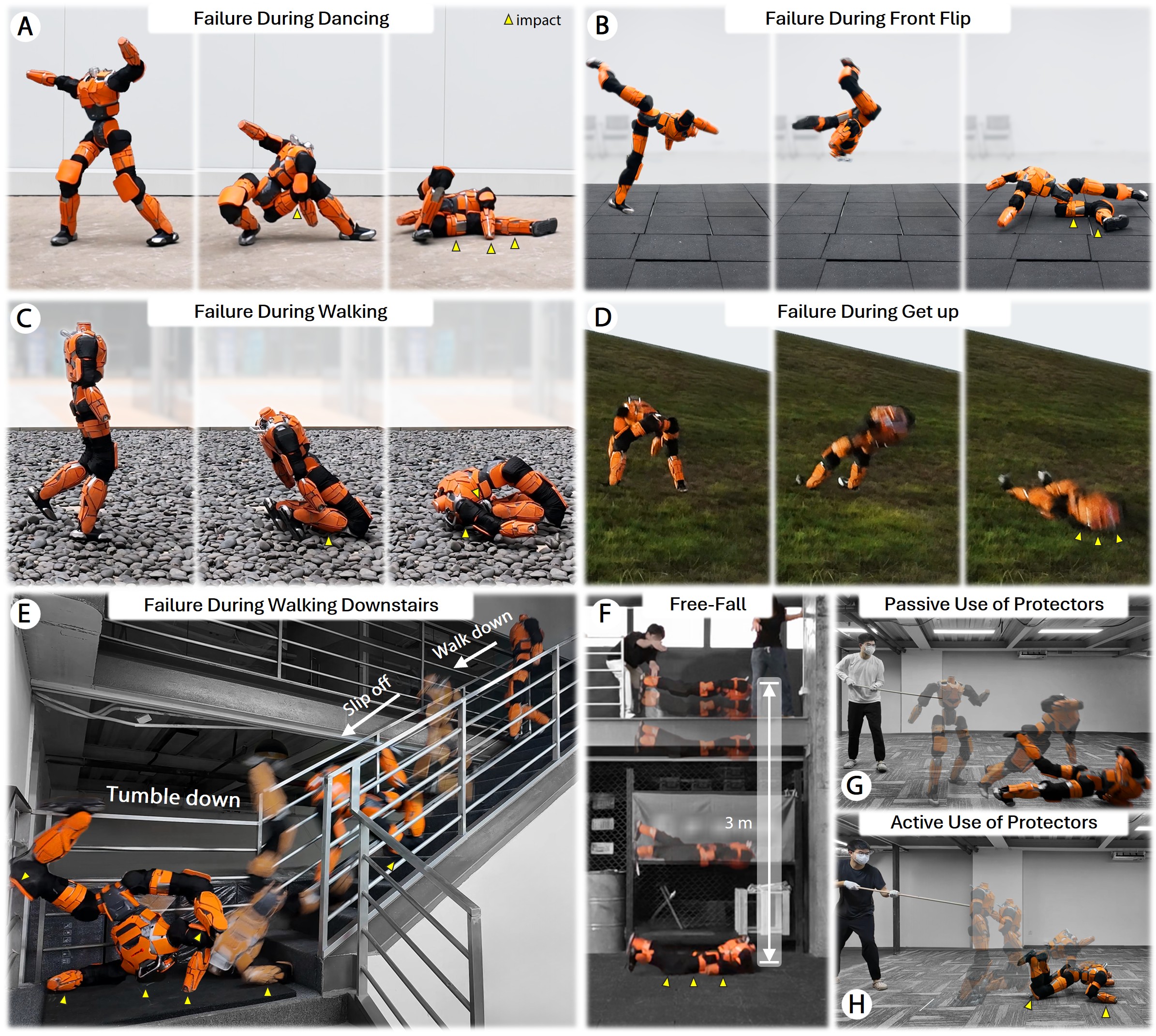} 

	\caption{\textbf{High-stress indoor and outdoor robustness tests.}
		(A) Power loss during dynamic dancing motions.
        (B) Mid-air power loss during a forward flip.
        (C) Power loss while walking on an outdoor cobblestone field.
        (D) Power loss during recovery on a steep outdoor slope.
        (E) Slip-induced tumbling during stair descent.
        (F) Fall from a second-floor drop.
        (G) Passive use of protectors: robot pushed over during dancing.
        (H) Active use of protectors: knee-first landing strategy enabled by thicker knee protectors during a push-over event.}
	\label{fig:real-highlight} 
\end{figure}

Beyond passive protection, we evaluated active fall strategies using control policies trained to exploit the deformable design during impact. Conventional vanilla reinforcement learning (RL) methods often failed to converge to stable solutions due to the large exploration space, complex reward functions, and discontinuous contact dynamics involved in humanoid falling. In contrast, with a mimic-based RL framework, the humanoid successfully learned coordinated falling motions that reoriented its body to ensure protected regions made first contact with the ground, thereby reducing the load on vulnerable joints (Fig.~\ref{fig:real-highlight}G-H).

Sequential motion recordings (Movie~S7) further revealed smoother deceleration profiles and shorter recovery times, with the robot effectively using its knees and upper limbs to dissipate energy during forward falls. Across repeated trials, the humanoid consistently executed active safe-fall behaviors without structural damage, confirming that active control can complement passive protection to enhance safety and resilience.

While these results demonstrate the potential of learning-based active fall control, achieving robust generalization across diverse scenarios will require larger and more varied demonstration datasets. Integrating this framework with physics-based simulation and reinforcement learning at scale represents an exciting direction for future work toward fully adaptive humanoid safety control.

\section*{Discussion}

This work presents a soft–rigid co-design framework that enables humanoid robots to interact safely and resiliently with their environment, especially during unplanned falls. By embedding a non-Newtonian soft responsive material (SRM) into rigid structures, the robot achieves two distinct mechanical behaviors: it remains compliant during normal operation and stiffens rapidly under impact. This property allows the surface to stay soft and adaptive in everyday motion while effectively absorbing and spreading impact energy when collisions occur. The protector meets the CE certification requirements for fall safety; furthermore, we perform quantitative calibration of the simulation model using the measured impact force–time curves and pressure distributions for protector design. This enables the model to reliably reproduce the experimental pressure fields in both magnitude and spatial pattern, achieving consistent agreement between simulation and experiment. Through three stages of validation, from material characterization to joint-level testing and full-body experiments, the results show consistent improvements in safety and robustness without noticeable loss in agility, heat dissipation, or reachability. The learning-based control policy further enhances protection by guiding the robot’s motion to direct impacts toward the protected areas, reducing structural stress during falls.

These findings address one of the most persistent challenges in humanoid robotics: ensuring physical safety without sacrificing performance. Conventional designs often rely on heavier rigid structures to withstand impacts, which increases inertia and amplifies the damage in actual collisions to surrounding environments. In contrast, the proposed design achieves comparable or superior protection using a lightweight and compact configuration. 
The approach can be generalized across different humanoid platforms through simulation-guided optimization of material thickness and placement, making it suitable for both research and industrial purposes. This flexibility allows the method to be integrated into future humanoid development pipelines as a standard design consideration for mechanical safety.

Beyond the immediate experimental results, the study provides broader insight into the design of physically safe robots. The effectiveness of the hybrid structure reflects principles observed in biological systems, where rigid bones provide mechanical strength and soft tissues cushion external impact. Translating this concept into humanoid design through advanced materials and adaptive control offers a realistic pathway toward safe, human-compatible robots that can operate reliably in unstructured environments.

While the proposed system shows strong performance, several aspects require further study. First, although extensive tests were conducted, long-term durability under material aging has not yet been characterized. The performance of the SRM composite may vary with temperature, humidity, and prolonged cyclic loading. Second, the current prototype focuses on surface-level protection. Deeper structural co-design, where compliant materials are integrated directly into load-bearing components, could further reduce weight and internal stress. Third, the simulation model remains a simplified proxy of the non-Newtonian material, capturing its macroscopic response but not the underlying particle-level mechanics, so more accurate modeling will be needed for complex collision scenarios. Finally, perception modules such as cameras and LiDAR sensors remain exposed. Future work will explore protective strategies for these components, balancing impact resistance with visual and sensing performance.


Overall, this research demonstrates that integrating responsive materials with intelligent control provides a practical path toward safer humanoid systems. The proposed framework achieves both hardware resilience and interaction safety, marking a step forward toward robust and human-compatible humanoids capable of reliable operation in everyday environments.


\section*{MATERIALS AND METHODS}



\subsection*{Material-level impact testing}

We evaluated the soft responsive material (SRM) following the European CE impact-protection standard~\cite{EN1621_1_2012}, widely used for certifying protective gear in motorcycle and outdoor sports. Because humanoid robots operate at similar mass scales and encounter comparable collision energies, this protocol provides a rigorous and relevant benchmark. An $5 \mathrm{kg}$ mass was dropped from a height of $1 \mathrm{m}$ onto each material sample, and the resulting contact forces were recorded at $25 \mathrm{kHz}$ to capture the rapid dynamics of impact shocks. These measurements were used both to assess impulse absorption performance and to calibrate the simulation model described later. Details of the testing apparatus are provided in the Supplementary Materials under "Material impact testing platform" section.

\subsection*{Joint-level pressure measurement}  

To characterize pressure distribution during joint impacts, we performed experiments on representative humanoid joint modules. Traditional electrical force sensors provide only point measurements or limited spatial coverage~\cite{ati}, and larger pad-type sensors often operate at insufficient temporal resolution ($50$–$250,\mathrm{Hz}$) due to processing and data-transfer delays~\cite{leboutet2019tactile, luo2024adaptive}. These limitations make them unsuitable for capturing sub-millisecond impact peaks typical of humanoid falls.

To obtain full-field, high-resolution measurements, we used Fujifilm Prescale pressure measurement film (PMF)~\cite{fujifilm}. This thin, flexible medium conforms to curved surfaces and permanently records peak pressure through color intensity, enabling accurate visualization of contact patterns and stress concentration.

We tested three representative humanoid joint configurations: single-degree-of-freedom (DoF) joints such as elbows and knees, two-DoF joints such as shoulders, and three-DoF joints such as hips. Each module was covered with medium-pressure MS PMF (see Supplementary Material under "Pressure measurement film" section) and subjected to repeated drop tests from a height of $1 \mathrm{m}$, both with and without the protective material. For each configuration, we analyzed (1) pressure distribution, (2) post-impact joint mobility and structural integrity, and (3) damage to the surrounding environment.

\subsection*{Full-body robot testing}  
In the final phase, we performed whole-body humanoid experiments to evaluate fall safety in realistic scenarios. The tests included common causes of failure such as power interruptions, external pushes, ground slips, and obstacle collisions. Medium-pressure MS PMF sheets were placed across the humanoid’s surface to capture distributed pressure during impact. Falls were repeated more than thirty times for the protected robot to obtain statistically reliable results. In contrast, unprotected robots typically experienced catastrophic hardware failure within the first three consecutive trials, requiring structural or electrical board repair before further testing.

\subsection*{RT-FEM analysis of soft responsive materials}
To characterize the dynamic behavior of the soft responsive material (SRM), we conducted controlled impact tests using three fixtures with different contact geometries, including hemisphere, square, and triangular pyramid. These experiments provided impulse–deformation data from which we extracted effective elastic modulus, rate-dependent stiffening behavior, and stress–strain hysteresis under high-speed loading. Based on these measurements, we built a fast real-time finite-element model (RT-FEM) that approximates the SRM’s dual behavior, being soft under low strain and significantly stiffer during rapid compression~\cite{fem-pbd}.

For design purposes, the simulation must be both accurate and efficient. We adopt a GPU-accelerated hyperelastic simulator~\cite{zeng2025fba} that incorporates nonlinear stiffness, rate dependence, and accurate contact handling. The solver uses a constraint-based formulation to robustly capture impact interactions and supports large batches of impact cases with stable convergence~\cite{xpbd}.

The full FEM pipeline includes collision detection, assembly of constraint terms, and an iterative local–global solve before updating positions and velocities. Thanks to this formulation, simulations run in real time or near real time, making it feasible to explore thousands of candidate protector designs. This capability allows thickness, geometry, and placement of SRM pads to be optimized directly from simulated impact forces, supporting rapid iteration and guiding the physical co-design process. Additional implementation details are provided in the Supplementary Materials under "Real-time FEM simulation framework" section.

\subsection*{Simulation framework for fall dynamics analysis}

We conducted large-scale fall simulations in MuJoCo~\cite{todorov2012mujoco} using the same locomotion policy as deployed on the physical humanoid robot. Four representative falling scenarios were simulated, external push, power loss, ground slip, and obstacle collision, as defined in the earlier sections. During each simulated episode, contact events, impact forces, and positions were recorded for all robot links to construct a comprehensive dataset of fall dynamics (Fig.~\ref{fig:mujuco-sample}F1).

A fall event was automatically detected when any link other than the two feet made contact with the ground. The accumulated data were used to analyze contact distributions, identify high-stress regions, and guide the placement and thickness of the protective materials in subsequent co-design stages (Fig.~\ref{fig:mujuco-sample}H).


\subsection*{Computational design and fabrication of 3D joint module protectors}

To determine the appropriate placement and thickness of the soft responsive material (SRM), we used simulation-guided analysis informed by fall dynamics. First, large-scale fall simulations estimated the peak contact forces at each link. These forces were then used as inputs to the real-time FEM (RT-FEM) analysis, which approximates the impact event and predicts the resulting pressure distribution across materials of different thicknesses. By iteratively adjusting the SRM thickness in simulation, we selected the minimum thickness required to keep local pressures below the material-dependent safety threshold. This simulation-to-design workflow establishes a principled mapping from fall impact forces to protector geometry. 

Because both the SRM sheets and PMF films are manufactured as planar layers while humanoid joints are highly curved, we applied a UV-mapping technique~\cite{Levy2002LSCM} to unwrap the 3D joint geometry into distortion-minimized 2D templates. This allowed the protective material to be precisely shaped for each surface while preserving alignment with high-stress regions identified in simulation.

The fabrication process balanced surface conformity and manufacturability. Each joint was segmented into a set of laser-cut patches that could be applied with minimal wrinkling or overlap. In total, $208$ segments were generated across all joint modules (per module: elbow 21, shoulder 13, hip 28, knee 27, torso 30). Both the SRM protector and the pressure measurement films were fabricated from these same UV-mapped templates, ensuring tight surface fitting, consistent sensor placement, and accurate comparison between protected and unprotected conditions (Fig.~\ref{fig:pipeline}C5).


\subsection*{Policy learning for active fall safety}

Beyond passive protection, the soft–rigid design enables the learning of active control policies for fall mitigation. The learning pipeline consists of two main stages:
(1) Motion capture and retargeting: expert demonstrations of safe, active falls were processed to generate reference trajectories; and
(2) Reinforcement learning (RL)–based motion mimicking: a policy was trained in simulation to reproduce and optimize these motions.

During training, the RL agent was guided with expert priors and optimized to minimize joint impacts during collisions, promoting smooth, energy-dissipating behaviors upon contact. The learned policy generated coordinated full-body movements that actively reoriented the robot to ensure that protected regions made first contact with the ground, effectively reducing impact magnitude compared with passive falling.

Compared with standard RL approaches that lacked expert prior knowledge, the mimic-based framework achieved more stable convergence and higher performance for this challenging problem~\cite{liao2025beyondmimic, he2025asap, tessler2024protomotions}, which involves large exploration spaces, complex reward structures, and discontinuous contact dynamics. The full training setup and reward definitions are provided in the Supplementary Materials under "Policy learning for active falling" section.

\clearpage 

%
\bibliography{science_template} 
\bibliographystyle{sciencemag}

%
%
%
%
%
%


\section*{Acknowledgments}
All authors thank the support of the material from Jiequn Wang, the discussion with Gang Yang, Zhenhao Huang, Yuhang Zheng, Zihao Xu, Tong Niu, Dongming Han, Yiran Liu, Xiaosheng Lin, and all members of ENGINEAI for their support. Artificial intelligence tools (ChatGPT) were used solely to assist with language refinement after the authors had completed the full initial draft. All scientific content, interpretations, and conclusions were generated by the authors. The final manuscript was thoroughly reviewed and verified by all authors to ensure accuracy and clarity.

\paragraph*{Funding:}
C.W, Z.Q.Z, Z.Z.Z, S.L and F.S. are supported in part by MOE 24-1234-P0001 and NUS Presidential Young Professorship. Y.Z and C.L are supported by the National Research Foundation (NRF), Prime Minister's Office, Singapore under its Campus for Research Excellence and Technological Enterprise (CREATE) programme. The Mens, Manus, and Machina (M3S) is an interdisciplinary research group (IRG) of the Singapore MIT Alliance for Research and Technology (SMART) centre, one of the research entity under the CREATE program. Q.G. is financially supported by the SNSF under project No. 197017. 
\paragraph*{Author contributions:}
Protector hardware development: C.W, Y.Z, A.T, H.C, Q.G, and F.S. Robot algorithm development: C.W, A.T, H.C, Z.Z.Z, B.L, Z.X, A.Z, C.H, T.Z, and F.S. Material simulation development: C.W, Y.Z, Z.Q.Z, H.C, Q.G, Z.Z.Z, A.Z, H.C, S.L, C.L, and F.S. Project management: C.L and F.S.
\paragraph*{Competing interests:}
There are no competing interests to declare.
\paragraph*{Data and materials availability:}
All data needed to evaluate the conclusions in the paper are present in the paper and the Supplementary Materials. Codes will be open-sourced.





\subsection*{Supplementary materials}
Materials and Methods\\
Figs. S1 to S7\\
References \textit{(44-\arabic{enumiv})}\\ 
Legends for movies S1 to S7\\
Movie S1 to S7\\

\newpage


\renewcommand{\thefigure}{S\arabic{figure}}
\renewcommand{\thetable}{S\arabic{table}}
\renewcommand{\theequation}{S\arabic{equation}}
\renewcommand{\thepage}{S\arabic{page}}
\setcounter{figure}{0}
\setcounter{table}{0}
\setcounter{equation}{0}
\setcounter{page}{1} 


\begin{center}
\section*{Supplementary Materials for\\ \scititle}

	Chunzheng~Wang$^{1\ast}$,
	Yiyuan~Zhang$^{1\ast}$,
	Annan~Tang$^{2\ast}$,
	Ziqiu~Zeng$^{1\ast}$,
    \\
	Haoran~Chen$^{1}$,
	Quan~Gao$^{3}$,
	Zixuan~Zhuang$^{1}$,
	Boyu~Li$^{2}$,
	Zhilin~Xiong$^{2}$,
    \\
	Aoqian~Zhang$^{1}$,
	Ce~Hao$^{1}$,
	Siyuan~Luo$^{1}$,
	Tongyang~Zhao$^{2}$,
    \\
	Cecilia~Laschi$^{1}$,
	Fan~Shi$^{1}$\\
	\small$^{1}$College of Design and Engineering, National University of Singapore, Singapore.\\
	\small$^{2}$ENGINEAI Robotics Technology Co., Ltd, China.\\
	\small$^{3}$Institute of Robotics and Intelligent Systems, ETH Zürich, Switzerland.\\
	\small$^\ast$These authors contributed equally to this work.
\end{center}

\subsubsection*{This PDF file includes:}
Materials and Methods\\
Figures S1 to S4\\
Legends for movies S1 to S7\\

\subsubsection*{Other Supplementary Materials for this manuscript:}
Movies S1 to S7\\

\newpage


\subsection*{Materials and Methods}

\subsection*{Pressure measurement film}

Pressure measurement films (PFMs) from Fujifilm~\cite{fujifilm} were used to quantify peak contact pressures during joint-level and whole-body impact experiments. PFMs consist of two paper-like layers coated with microcapsules that rupture under load; the resulting red coloration encodes the peak pressure distribution, where darker regions correspond to higher pressures and lighter regions correspond to lower loads. Pressure values can be read either through a calibrated color reference chart or obtained digitally using analysis software that converts the color intensity into spatial pressure maps.

Because the pressure ranges encountered in humanoid falling vary widely across scenarios, multiple film types with different measurement ranges were employed.
\begin{itemize}
    \item Medium-pressure (MS) films (lower bound $10$ MPa): used on the robot body to evaluate pressure distributions during falling tests and to characterize material behavior.
    \item Low-pressure (LW) films, (lower bound $2.5$ MPa): used for low-magnitude interactions, such as arm collisions with a punching bag.
    \item  High-pressure (HS) films, (lower bound $50$ MPa): used for humanoids controlled pushing-down tests onto the ground, where local impact loads are substantially higher.
\end{itemize}

These PFMs provide a thin, flexible, and high-resolution sensing modality, enabling accurate measurement of pressure distributions without altering the robot’s geometry or interfering with motion.

\subsection*{Humanoid robot platform}

\begin{figure} 
	\centering
	\includegraphics[width=1\textwidth, trim=0 0 0 160, clip]{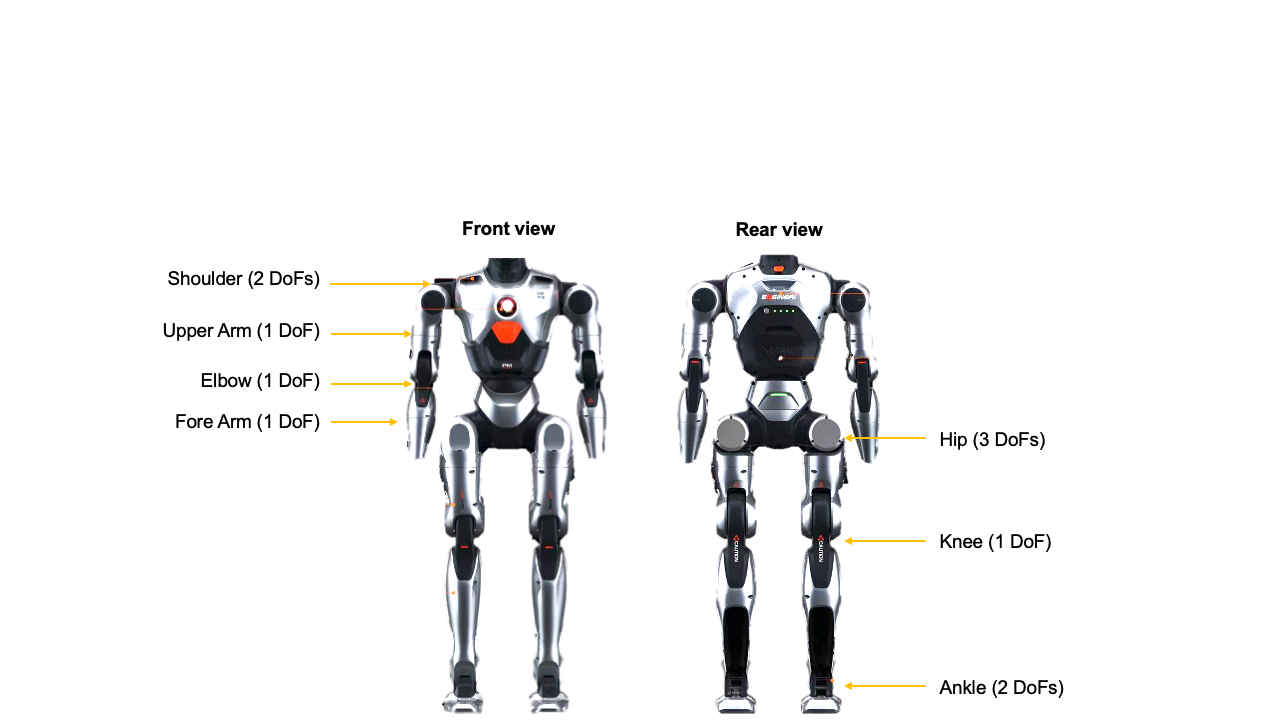}
	\caption{{\textbf{The tested humanoid robot: ENGINEAI PM01.} }}
	\label{fig:supp-robot} 
\end{figure}

The humanoid robot used in our experiments is the life-size ENGINEAI PM$01$~\cite{engineai}, which stands $1.38$ m tall and weighs $42$ kg (including battery). The robot is equipped with $22$ actuated joints: $5$ motors per arm and $6$ per leg. Two motor types are used, providing peak torques of $145$ Nm and $50$ Nm, respectively. PM$01$ is a powerful platform capable of dynamic motions such as running, dancing, and front flips.

Although PM$01$ serves as our demonstration platform, the proposed method is general and applicable to a wide range of humanoid robot designs.

\subsection*{Material impact testing platform}

\begin{figure} 
	\centering
	\includegraphics[width=1\textwidth, trim=0 0 0 180, clip]{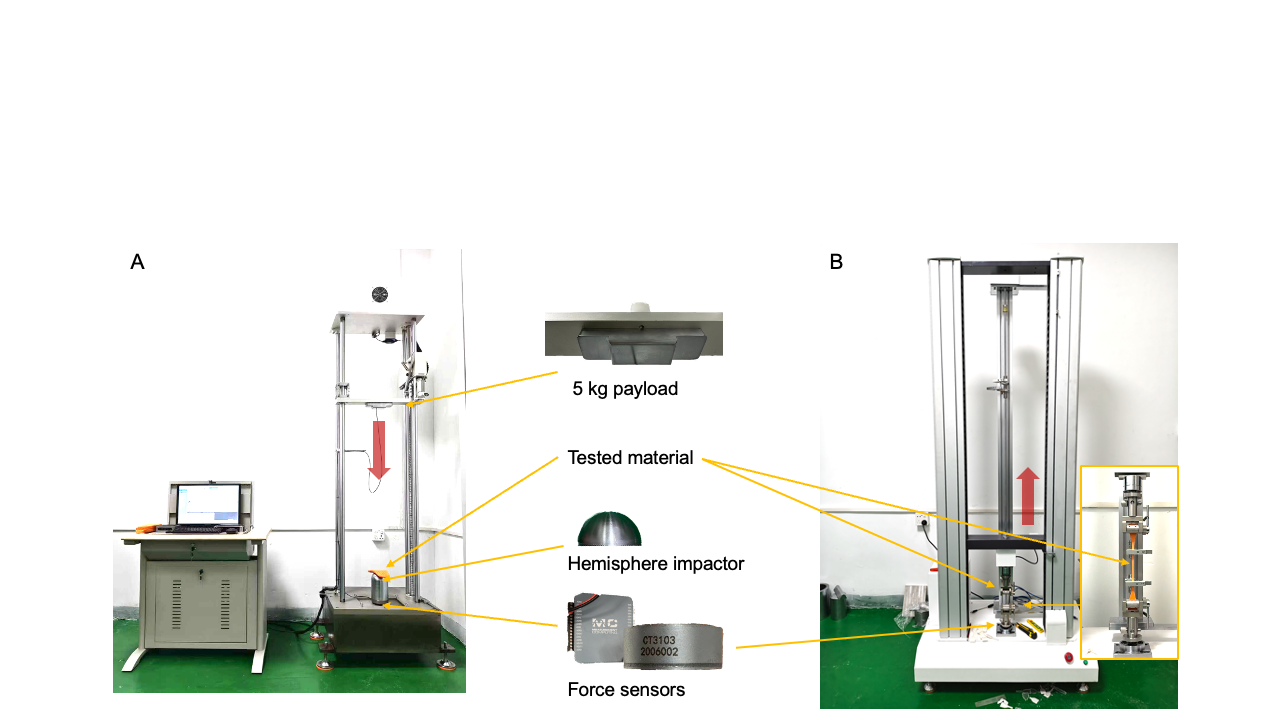} 

	\caption{{\textbf{Material characterization platform.}
        (A) CE-standard drop-test device.
        (B) Stretching machine for strain and tensile testing.}}
	\label{fig:supp-ce-testbed} 
\end{figure}

Material-level impact performance was evaluated using a CE-standard drop-test device (Fig.~\ref{fig:supp-ce-testbed} following the EN1621-1 protocol~\cite{EN1621_1_2012}. The setup consists of a rigid payload ($5$ kg) with a hemispherical impactor at its base, which is released in free fall from a height of $1$ m onto the test sample placed on a flat support surface. A high-frequency force sensor is mounted beneath the sample to capture the full force–time profile of the impact, enabling accurate measurement of peak force and impulse characteristics. This platform provides a controlled and repeatable environment for comparing different materials under standardized impact conditions.

\subsection*{Real-time FEM simulation framework}

Our simulator builds on the real-time GPU-based hyperelastic FEM framework introduced in \cite{zeng2025fba}. The method is designed for scenarios involving large deformations, dense contact, and friction, conditions typical of humanoid impacts, while still achieving real-time or near–real-time performance. These settings are challenging because they involve strong material nonlinearities and non-smooth contact constraints, which traditionally make implicit FEM solvers slow or unstable. For our co-design pipeline, which requires thousands of evaluations, computational efficiency is essential.

Our RT-FEM framework reformulates the classical local–global method into a GPU-parallel implicit solver. Contact constraints, including friction, are embedded directly into the local–global updates through a complementarity formulation, allowing elastic forces and contact forces to be solved within a unified system. The global step is reduced to sparse matrix multiplications, enabling efficient parallelization and avoiding repeated factorization.

To further improve scalability, the solver uses a compact representation of the system inverse and a splitting strategy for non-smooth constraints, which stabilizes convergence in cases with dense or high-friction contact. These features allow the method to remain robust across a broad range of hyperelastic materials, from highly compliant to extremely stiff, and under impact conditions with numerous simultaneous contact points.

Performance benchmarks reported in \cite{zeng2025fba} show that the solver typically runs below 30 ms per timestep for meshes exceeding $10,000$ vertices, even under extreme deformation, stick–slip transitions, and multi-contact impacts. All benchmarks were conducted on a workstation with an Intel i9-13900KF CPU and an NVIDIA RTX 4090 GPU.

\begin{figure} 
	\centering
	\includegraphics[width=0.9\textwidth, trim=0 0 0 140, clip]{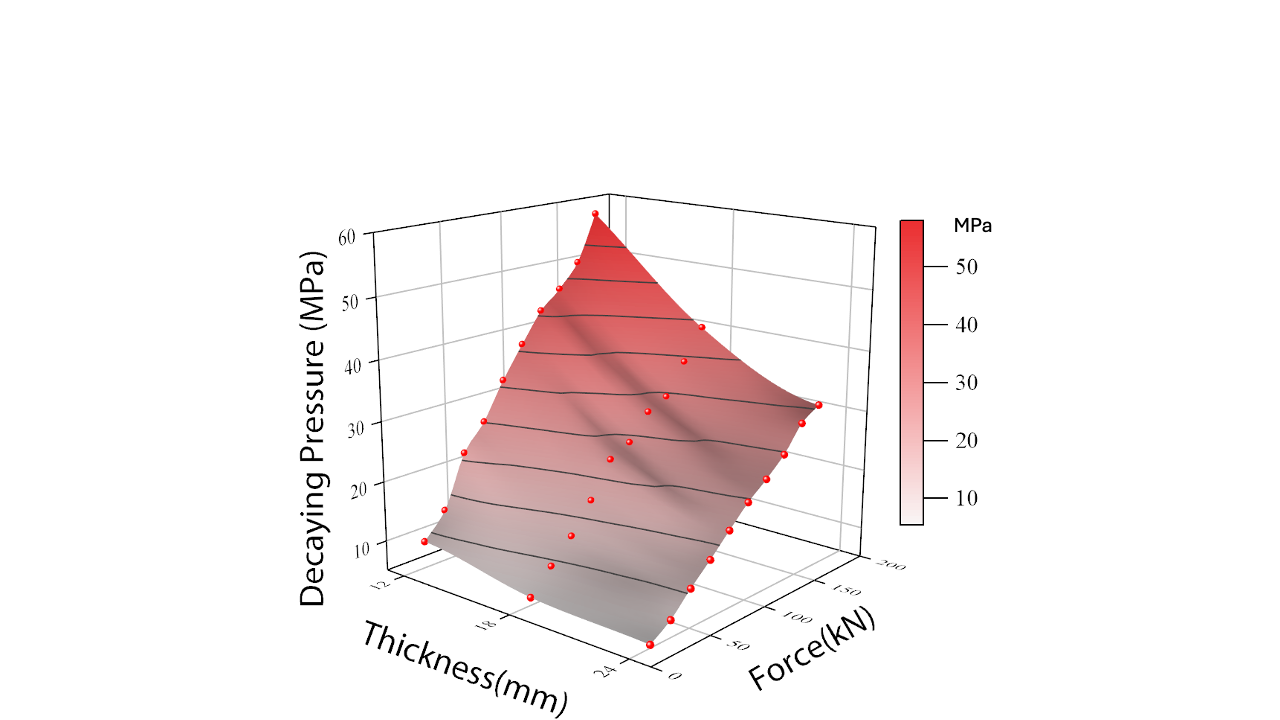} 

	\caption{{\textbf{RT-FEM results for optimizing SRM thickness.}
        The real-time FEM analysis quantifies how varying SRM thickness reduces peak pressure across a range of impact forces.
        }}
	\label{fig:supp-sim-results} 
\end{figure}

For our application, this real-time capability enables large-scale simulation-in-the-loop design: rapid evaluation of protector geometry and thickness variations (Fig.~\ref{fig:supp-sim-results}), high-throughput impact sampling, and iterative refinement based on simulated pressure and force predictions. Such throughput would not be feasible using conventional FEM pipelines.

\subsection*{Policy learning for active falling}

\textbf{Vanilla RL baseline.} 
We first attempted to train a reinforcement learning (RL) policy using conventional reward shaping without motion priors. However, the training failed to converge to meaningful fall mitigation behaviors. The difficulty arises from the high dimensionality of humanoid control and the discontinuous contact dynamics inherent in falling, which make it challenging to encode desired motion patterns through simple fixed rewards. These observations motivated the use of a mimic-based learning approach that incorporates expert demonstrations to provide structured guidance and improve policy feasibility on the humanoid robot.

\begin{figure} 
	\centering
	\includegraphics[width=1\textwidth]{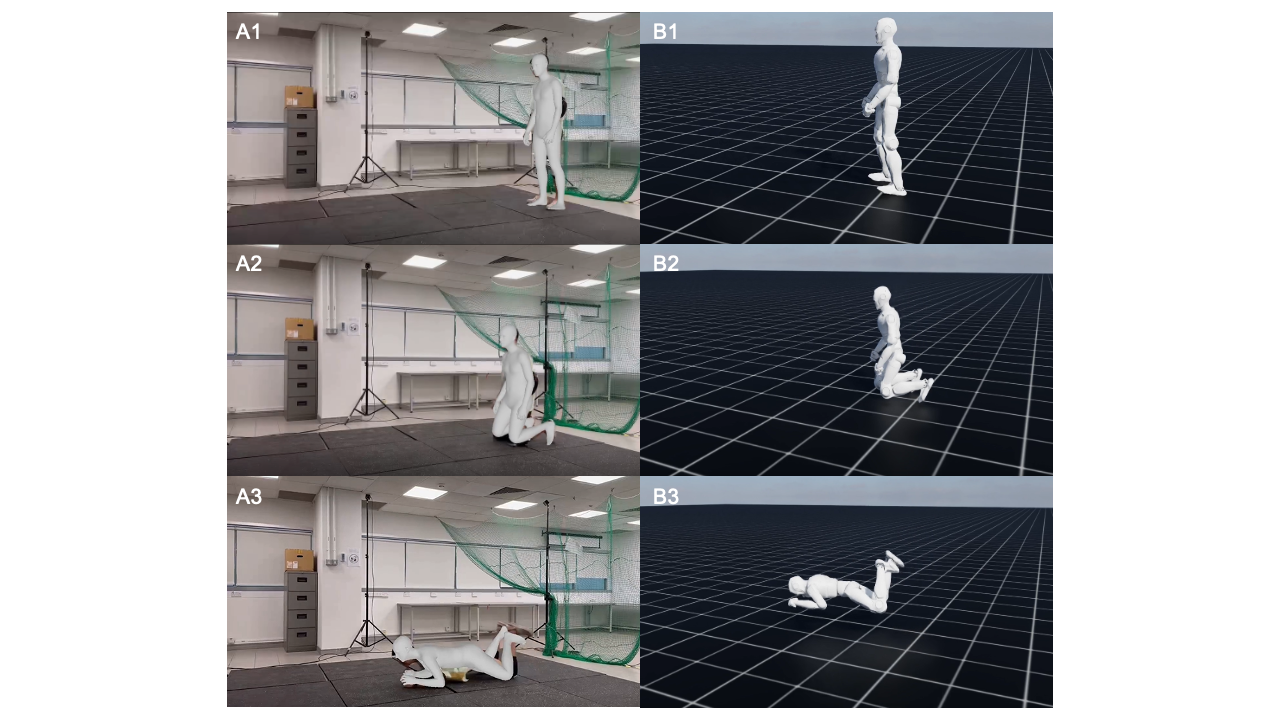} 

	\caption{{\textbf{Retargeting human expert video demonstrations to the robot configuration.}
        (A1–A3) Reconstruction of 3D human expert motion from a monocular video sequence.
        (B1–B3) Retargeted motion mapped to the humanoid robot configuration in simulation, used as a prior for policy learning.}}
	\label{fig:supp-retarget} 
\end{figure}

\textbf{Stage 1 – Motion capture and retargeting.}
This stage provides high-quality reference data for policy learning. A robot expert wearing the same protective material as the humanoid acted as a proxy to demonstrate active, controlled falling motions. The movements were recorded using a monocular RGB camera at $30$ Hz. We employed GVHMR~\cite{shen2024world} to reconstruct 3D human motion in the SMPL format from these videos as Fig.~\ref{fig:supp-retarget} (A1-A3). The reconstructed motions were then retargeted to our life-size humanoid model using GMR~\cite{araujo2025retargeting}, and interpolated to $50$ Hz to achieve smoother and more precise control as Fig.~\ref{fig:supp-retarget} (B1-B3).

Compared with other retargeting approaches such as PHC~\cite{luo2023perpetual,he2024learning,he2025asap,he2024omnih2o} and ProtoMotions~\cite{tessler2024protomotions}, GMR produced more stable contact sequences and reduced artifacts such as body floating and foot sliding. The output of this stage is a dataset of physically plausible, retargeted fall trajectories $\tau_{ref}$, which serve as reference motions for policy learning.

\textbf{Stage 2 – RL-based motion mimicking.}
We formulate motion mimicking as a reinforcement learning problem, enabling the robot to adapt the reference trajectory $\tau_{ref} = (q_m, v_m)$ to its embodiment and dynamics. To maintain fall style while mitigating global drift, an issue common in dynamic contact-rich tasks, we define tracking objectives relative to an anchor body $b_{anchor}$ (e.g., the pelvis or torso). The desired pose $\hat{T}_{b_{anchor}}$ tracks the reference motion directly, while all other target bodies $b \in \mathcal{B}_{target}$ track the reference motion relative to the current anchor state. This hybrid coordinate strategy aligns reference motion height and orientation with the simulated robot, preventing cumulative drift. The desired body twists $\hat{\mathcal{V}}_b$ are set equal to the reference motion velocities $\mathcal{V}_{b,m}$.

\textbf{Observations.}
The policy observation vector includes:
(1) Reference phase: joint positions and velocities from the reference motion $[q_{joint,m}, v_{joint,m}]$, providing phase context;
(2) Anchor pose-tracking error: $\xi_{b_{anchor}} \in \mathbb{R}^9$, consisting of 3D position and rotation errors;
(3) Proprioception: the robot’s current root twist (in the root frame), joint states $(q_{joint}, v_{joint})$, and the previous action $a_{last}$.

\textbf{Rewards.}
The total reward $r$ encourages motion fidelity, safe impact behavior, and smooth control, while penalizing instability. The primary task reward $r_{task}$ tracks reference poses and velocities using normalized exponential penalties:
\begin{equation}
    r(\overline{e}_{\chi}) = \exp(-\overline{e}_{\chi} / \sigma_{\chi}^2), \hspace{5mm} r_{task} = \sum_{\chi \in \{p, R, v, w\}} r(\overline{e}_{\chi})
\end{equation}
where $\overline{e}_{\chi}$ are position, rotation, and velocity errors, and $\sigma_{\chi}$ are nominal scaling factors.

To promote active safety, a safe-contact reward $r_{safe}$ provides positive feedback when designated body parts $b \in \mathcal{B}_{safe}$ (e.g., forearms) contact the ground for fall mitigation:
\begin{equation}
    r_{safe} = \sum_{b \in \mathcal{B}_{safe}} w_b \cdot \mathbb{I}(b \text{ is in contact with ground})
\end{equation}
where $w_b$ is a weighting coefficient.

The total reward is:
\begin{equation}
r = \lambda_{task} r_{task} + \lambda_{safe} r_{safe} + \lambda_l r_{limit} + \lambda_s r_{smooth} + \lambda_c r_{contact}
\end{equation}
where $r_{limit}$, $r_{smooth}$, and $r_{contact}$ penalize joint acceleration, jerky actions, and self-collisions, respectively.

\textbf{Termination and sampling.}
Training episodes terminate when: (1) the anchor body’s height or orientation errors exceed predefined thresholds, or (2) critical body parts such as the head make unsafe contact with the ground. At each reset, the robot’s state is initialized to a sampled phase from $\tau_{ref}$ with added random perturbations to improve robustness.

To prioritize difficult fall phases, we adapt the sampling strategy from BeyondMimic~\cite{liao2025beyondmimic}. The trajectory $\tau_{ref}$ is divided into temporal bins, and sampling probabilities are weighted by recent failure rates, ensuring efficient training across challenging fall segments.

\textbf{Domain randomization.}
To bridge the sim-to-real gap, we apply domain randomization over key simulation parameters, including ground friction, nominal joint positions $\overline{q}_j$ (representing calibration error), and the torso’s center of mass. Random external perturbations are also applied during training to encourage robust recovery behaviors.

\section*{Supplementary Movies}
\textbf{Supplementary Movie S1: }Overview of soft responsive materials for humanoid safety. Corresponds to Fig.~\ref{fig:pipeline}. Demonstrates the motivation, main concepts, and key experimental results of the proposed framework.
\\
\textbf{Supplementary Movie S2: }Real-time FEM simulation of the soft responsive material. Corresponds to Fig.~\ref{fig:material}. Shows real-time impact simulations and their alignment with physical tests.
\\
\textbf{Supplementary Movie S3: }Joint-level protection tests. Corresponds to Fig.~\ref{fig:joint-design}. Presents repeated elbow-drop experiments illustrating the effectiveness of the SRM joint protectors.
\\
\textbf{Supplementary Movie S4: }Representative fall simulations for damage mapping. Corresponds to Fig.~\ref{fig:mujuco-sample}. Shows four common humanoid failure scenarios used to generate large-scale contact statistics. 
\\
\textbf{Supplementary Movie S5: }Comparison of multiple failure modes with and without protectors. Corresponds to Fig.~\ref{fig:pressure-compare}. Shows life-size humanoid falls across four common scenarios and the resulting PMF measurements for protected and unprotected conditions.
\\
\textbf{Supplementary Movie S6: }Environmental safety evaluation. Corresponds to Fig.~\ref{fig:env-protect}. Shows ground-impact and punching-bag tests, highlighting the reduced environmental loading achieved with SRM protection.
\\
\textbf{Supplementary Movie S7: }High-intensity indoor and outdoor protection experiments. Corresponds to Fig.~\ref{fig:real-highlight}. Demonstrates the humanoid remaining safe and fully functional under challenging scenarios, including multi-level drops, stair descents, uneven outdoor terrain, and dynamic motions such as running, turning, and front flipping.
\\


\clearpage 



\end{document}